\PassOptionsToPackage{margin=1in}{geometry}
\documentclass{article}

\usepackage[preprint]{neurips_2025}
\usepackage{amsmath}
\usepackage{longtable}
\usepackage{listings}
\usepackage{float}
\usepackage{amssymb}
\usepackage{multicol}
\usepackage{blindtext}
\usepackage{tabularx}

\usepackage{multirow} 

\usepackage[utf8]{inputenc} 
\usepackage[T1]{fontenc}    
\usepackage{hyperref}       
\usepackage{url}            
\usepackage{booktabs}       
\usepackage{amsfonts}       
\usepackage{nicefrac}       
\usepackage{microtype}      
\usepackage{xcolor}         
\usepackage{subcaption}
\usepackage{graphicx}
\usepackage{tikz}
\usepackage{enumitem}
\usepackage{fontawesome5}
\usepackage{array}

\usepackage[T1]{fontenc}
\usetikzlibrary{shadows,fit,shapes.callouts,positioning}
\usetikzlibrary{positioning,shapes.geometric,arrows.meta}
\usetikzlibrary{backgrounds}
\tikzstyle{startstop} = [rectangle, rounded corners, minimum width=3cm, minimum height=1cm,text centered, draw=black, fill=white]
\tikzstyle{process} = [rectangle, minimum width=3cm, minimum height=1cm, text centered, draw=black, fill=white]
\tikzstyle{decision} = [diamond, aspect=2, minimum width=3cm, minimum height=1cm, text centered, draw=black, fill=white]
\tikzstyle{arrow} = [thick,->,>=stealth]
\title{Talking with Oompa Loompas: A novel framework for evaluating linguistic acquisition of LLM agents}

\author{%
  Sankalp Tattwadarshi Swain\thanks{Equal Contributions} \\
  BITS Pilani, India\\
  \texttt{f20230769@pilani.bits-pilani.ac.in} \\
  \And
  Anshika Krishnatray\footnotemark[1] \\
  BITS Pilani, India\\
  \texttt{f20230610@pilani.bits-pilani.ac.in} \\
  \And
  Dhruv Kumar \\
  BITS Pilani, India \\
  \texttt{dhruv.kumar@pilani.bits-pilani.ac.in} \\
  \And
  Jagat Sesh Challa \\
  BITS Pilani, India \\
  \texttt{jagatsesh@pilani.bits-pilani.ac.in} \\
}

\begin{document}

\maketitle

\begin{abstract}
Existing evaluation studies on linguistic competence of large language models (LLM agents) have focused primarily on vocabulary learning, morphological rule induction, syntactic generalization, pragmatic inference, and cross-linguistic transfer. However, none assess whether LLM agents can acquire a language through pattern recognition and interactive feedback, a central feature of human language acquisition. We propose a novel experimental framework in which an LLM agent is evaluated on its ability to acquire and use a newly constructed language (Tinkatongue) in conversation with a bot that understands only Tinkatongue. Our findings show that LLM agents fail to establish a conversation within 100 responses, yet they adopt distinct strategies that mirror human approaches to language learning. The results suggest a new direction for evaluation benchmarks and open pathways to model designs that learn more effectively from interactive feedback.
\end{abstract}

\section{Introduction}
The central problem that motivates this study is to understand how LLM agents acquire entirely new and novel languages. While current benchmarks demonstrate model fluency in existing human languages, it remains unclear whether these results reflect true language acquisition or simple memorization of patterns seen during training. This raises a fundamental research question: \textit{can LLM agents develop proficiency in a constructed language through mechanisms similar to human second-language learning, namely by recognizing patterns and adapting through iterative interaction and feedback?} Addressing this question not only provides insight into the cognitive-like abilities of LLM agents but also helps clarify whether their performance stems from genuine generalization capabilities or from reliance on prior exposure. This work aims to shed light on the extent to which they can mimic human strategies of feedback-driven improvement for language acquisition.

Current evaluation methods for large language models (LLM agents) primarily focus on tasks within existing languages, such as classification, reasoning, memorization and cross-lingual transfer, using benchmarks like GLUE (\textit{\cite{Wang2018GLUEAM}}), SuperGLUE (\textit{\cite{Wang2019SuperGLUEAS}}) and MMLU (\textit{\cite{Hendrycks2020MeasuringMM}}). Studies on linguistic competence of LLM agentS have examined their abilities in vocabulary learning, morphological rule induction (\textit{\cite{Weissweiler2023CountingTB}}), syntactic generalization (\textit{\cite{hu2020systematicassessmentsyntacticgeneralization}}), and pragmatic inference (\textit{\cite{Park2024MultiPragEvalMP}}), with some research investigating cross-linguistic transfer (\textit{\cite{Artetxe2019OnTC}}). Methods such as
fine-tuning and prefix-tuning (\textit{\cite{Meloux2024NovelWDEA}}) have been explored to adapt LLM agents to specific domains or tasks. However, there remains a critical gap in understanding whether LLM agents can acquire entirely new languages through pattern recognition and interactive feedback during runtime. This gap highlights the need for novel evaluation approaches that test the ability of LLM agents to adapt to new linguistic systems in real time.

In this study, we propose a novel evaluation method to assess the ability of LLM agents to learn a new language through interaction. The LLM agent is tasked with conversing with a bot, \texttt{Oompa Loompa}, that understands only a newly constructed language, \texttt{Tinkatongue}. The LLM agent has no prior knowledge of \texttt{Tinkatongue}. The goal of the LLM agent is to communicate successfully with the bot, \texttt{Oompa Loompa},  by generating valid sentences in \texttt{Tinkatongue}. \texttt{Oompa Loompa} provides feedback to the LLM agent, indicating whether its response is valid or not based on a predefined set of syntactic rules. This method evaluates the LLM agents' ability to acquire a language dynamically, relying on pattern recognition and real-time feedback, simulating the human-like process of language acquisition through interaction.

Experimental results demonstrate that Claude-3.5-haiku consistently outperformed GPT-4o-mini and Gemini-2.5-flash across multiple metrics. All models demonstrated high Feedback Responsiveness, recovering well from mistakes once valid sentences were identified. Despite these improvements, no model achieved a fully successful conversation within 100 responses, highlighting the challenge of sustained language learning. Qualitative analysis revealed that the models used strategies such as imitation, babbling, and systematic combinatorial testing, which mirror stages in human language acquisition. These behaviors suggest that LLM agents adapt to new linguistic environments through feedback-driven exploration, providing insights into the potential of interactive language acquisition in artificial systems.

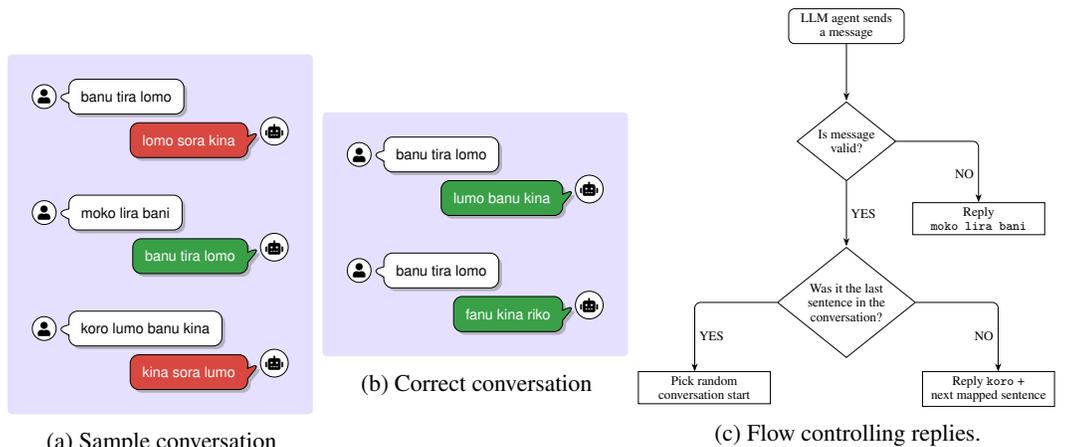
\begin{figure}[b]
  \centering
  \begin{subfigure}[b]{0.60\textwidth}
    \centering
    \begin{minipage}[c]{0.49\linewidth}
      \centering
      \resizebox{\linewidth}{!}{%
        \begin{tikzpicture}[font=\sffamily, every node/.style={font=\sffamily}, transform shape]
                \definecolor{good}{RGB}{57,160,72}
                \definecolor{bad}{RGB}{216,72,64}
                \definecolor{framepurple}{RGB}{120,86,255}
                \node[circle,draw,inner sep=1.2pt, fill = white] (h1) {\faUser};
                \node[rectangle callout, rounded corners=6pt, draw, fill=white,
                      align=left, inner sep=3mm, callout absolute pointer={(h1.east)},
                      right=3mm of h1, drop shadow] (hb1) {banu tira lomo};
                \node[circle,draw,inner sep=1.2pt, right=5cm of h1, yshift=-8.5mm, fill = white] (r1) {\faRobot};
                \node[rectangle callout, rounded corners=6pt, draw, fill=bad, text=white,
                      align=left, inner sep=3mm, callout absolute pointer={(r1.west)},
                      left=3mm of r1, yshift=-2.2mm, drop shadow] (rb1) {lomo sora kina};
                \node[circle,draw,inner sep=1.2pt, below=14mm of h1, yshift=-8.5mm, fill = white] (h2) {\faUser};
                \node[rectangle callout, rounded corners=6pt, draw, fill=white,
                      align=left, inner sep=3mm, callout absolute pointer={(h2.east)},
                      right=3mm of h2, drop shadow] (hb2) {moko lira bani};
                \node[circle,draw,inner sep=1.2pt, right=5cm of h2, yshift=-8.5mm, fill = white] (r2) {\faRobot}; 
                \node[rectangle callout, rounded corners=6pt, draw, fill=good, text=white,
                      align=left, inner sep=3mm, callout absolute pointer={(r2.west)},
                      left=3mm of r2, yshift=-2.2mm, drop shadow] (rb2) {banu tira lomo};
                \node[circle,draw,inner sep=1.2pt, below=14mm of h2, yshift=-8.5mm, fill = white] (h3) {\faUser};
                \node[rectangle callout, rounded corners=6pt, draw, fill=white,
                      align=left, inner sep=3mm, callout absolute pointer={(h3.east)},
                      right=3mm of h3, drop shadow] (hb3) {koro lumo banu kina};
                \node[circle,draw,inner sep=1.2pt, right=5cm of h3, yshift=-8.5mm, fill = white] (r3) {\faRobot};
                \node[rectangle callout, rounded corners=6pt, draw, fill=bad, text=white,
                      align=left, inner sep=3mm, callout absolute pointer={(r3.west)},
                      left=3mm of r3, yshift=-2.2mm, drop shadow] (rb3) {kina sora lumo};
                \begin{scope}[on background layer]
                  \node[
                    rounded corners=3pt, inner sep=6mm,
                    fill=framepurple!60,   
                    fill opacity=0.3,      
                    fit=(h1)(hb1)(rb1)(r1)(h2)(hb2)(rb2)(r2)(h3)(hb3)(rb3)(r3)
                  ] {};
                \end{scope}

              \end{tikzpicture}
      }
    
      \subcaption{Sample conversation}
      \label{fig:conv-incorrect}
    \end{minipage}\hfill
    \begin{minipage}[c]{0.49\linewidth}
      \centering
      \resizebox{\linewidth}{!}{%
         \begin{tikzpicture}[font=\sffamily, every node/.style={font=\sffamily}, transform shape]
            \definecolor{good}{RGB}{57,160,72}
            \definecolor{bad}{RGB}{216,72,64}
            \definecolor{framepurple}{RGB}{120,86,255}
            \node[circle,draw,inner sep=1.2pt, fill = white] (h1b) {\faUser};
            \node[rectangle callout, rounded corners=6pt, draw, fill=white,
                  align=left, inner sep=3mm, callout absolute pointer={(h1b.east)},
                  right=3mm of h1b, drop shadow] (hb1b) {banu tira lomo};
            \node[circle,draw,inner sep=1.2pt, right=5cm of h1b, yshift=-8.5mm, fill = white] (r1b) {\faRobot};
            \node[rectangle callout, rounded corners=6pt, draw, fill=good, text=white,
                  align=left, inner sep=3mm, callout absolute pointer={(r1b.west)},
                  left=3mm of r1b, yshift=-2.2mm, drop shadow] (rb1b) {lumo banu kina};
            \node[circle,draw,inner sep=1.2pt, below=14mm of h1b, yshift=-8.5mm, fill = white] (h2b) {\faUser};
            \node[rectangle callout, rounded corners=6pt, draw, fill=white,
                  align=left, inner sep=3mm, callout absolute pointer={(h2b.east)},
                  right=3mm of h2b, drop shadow] (hb2b) {banu tira lomo};
            \node[circle,draw,inner sep=1.2pt, right=5cm of h2b, yshift=-8.5mm, fill = white] (r2b) {\faRobot};
            \node[rectangle callout, rounded corners=6pt, draw, fill=good, text=white,
                  align=left, inner sep=3mm, callout absolute pointer={(r2b.west)},
                  left=3mm of r2b, yshift=-2.2mm, drop shadow] (rb2b) {fanu kina riko};
            \begin{scope}[on background layer]
              \node[
                rounded corners=3pt, inner sep=6mm,
                fill=framepurple!60,   
                fill opacity=0.3,      
                fit=(h1b)(hb1b)(rb1b)(r1b)(h2b)(hb2b)(rb2b)(r2b)
              ] {};
            \end{scope}
            
            
          \end{tikzpicture}
      }
      
      \subcaption{Correct conversation}
      \label{fig:conv-correct}
    \end{minipage}
    \vspace{-1mm}
  \end{subfigure}\hfill
  \begin{subfigure}[b]{0.4\textwidth}
    \centering
    \resizebox{\linewidth}{!}{%
        \begin{tikzpicture}[
          node distance=14mm and 24mm,
          >=Stealth,
          every node/.style={font=\small,inner sep=3pt},
          startstop/.style={rectangle,rounded corners,draw,minimum width=28mm,align=center},
          process/.style={rectangle,draw,align=center,minimum width=32mm},
          decision/.style={diamond,aspect=1.25,draw,align=center,inner sep=2pt,minimum width=20mm},
          shorten >=1pt
        ]
        
        \node (start)   [startstop] {LLM agent sends\\a message};
        \node (valid)   [decision, below=of start] {Is message\\valid?};
        \node (last)    [decision, below=of valid, yshift=-2mm] {Was it the last\\sentence in the\\conversation?};
        
        \node (replyb)  [process, below right=10mm and 10mm of valid] {Reply\\\texttt{moko lira bani}};
        \node (restart) [process, below left=10mm and 10mm of last] {Pick random\\conversation start};
        \node (replyk)  [process, below right=10mm and 10mm of last] {Reply \texttt{koro} +\\next mapped sentence};
        
        \draw[->,rounded corners] (start.south) -- (valid.north);
        \draw[->,rounded corners] (valid.south) -- node[midway,right]{YES} (last.north);

        \path (valid.east) ++(20mm,0) coordinate (bendNO1);
        \path (last.west)  ++(-20mm,0) coordinate (bendYES2);
        \path (last.east)  ++(20mm,0) coordinate (bendNO2);
        
        \draw[->,rounded corners] (valid.east) -- ++(20mm,0) -- ++(0,-8mm)
              -- ++(0,-7mm) node[pos=0.001,left]{NO}; 
        \draw[->,rounded corners] (last.west) -- ++(-20mm,0) -- ++(0,-8mm)
              -- ++(0,-9mm) node[pos=0.001,right]{YES}; 

        \draw[->,rounded corners] (last.east) -- ++(20mm,0) -- ++(0,-8mm)
              -- ++(0,-9mm) node[pos=0.001,left]{NO}; 
        \end{tikzpicture}
        
     }
    \subcaption{Flow controlling replies.}
    \label{fig:flow}
  \end{subfigure}

  \caption{Side-by-side comparison: (a), (b) -  conversation outcomes; (c) reply flow. Note: The white chat boxes are the Oompa Loompa's responses. Red and Green chats are LLM agent's responses. Red chat indicates an invalid sentence, and Green chat indicates a valid sentence}
  \label{fig:conversation-and-flow}
\end{figure}

\section{Methodology}

\textbf{Problem Statement.} We formalize the task as an interaction between a large language model (LLM agent) and a deterministic agent, \texttt{Oompa Loompa},  that speaks a newly constructed language, \texttt{Tinkatongue} (Formal Language Specification are mentioned in Appendix A), that the LLM agent is unaware of. The \texttt{Oompa Loompa} enforces the grammar of \texttt{Tinkatongue} and provides structured feedback to the interacting LLM agent. The objective of LLM agent is to minimize the expected number of turns required to complete a conversation under \texttt{Oompa Loompa}’s feedback mechanism. Mathematical definitions are mentioned in Appendix B.

\textbf{Experimental Setup.} The interaction loop begins with \texttt{Oompa Loompa} producing an initial utterance and alternates turns until the LLM agent completes a full conversation or a preconfigured round limit \(T_{\max}\) is reached. The deterministic behavior of \texttt{Oompa Loompa} isolates variability due to the model and supports reproducible measurement of evaluation metrics.

\textbf{Feedback Mechanism.} If the response from the LLM agent is valid and not final, the \texttt{Oompa Loompa} emits positive feedback("koro") concatenated with the next mapped sentence as shown in Fig. \ref{fig:conv-incorrect}. If the response completes the dialogue, the \texttt{Oompa Loompa} registers dialogue completion and samples a new conversation to start. If the response is invalid, the \texttt{Oompa Loompa} replies with a confused sentences (``moko lira bani'') which resets the conversation state, and terminates the attempt, a sample conversation of such type is depicted in Fig. \ref{fig:conv-correct}. An immediately subsequent valid reply by the LLM agent is recorded as an \emph{immediate recovery}.

\newcommand{\xgap}{5.2cm}   
\newcommand{\yrow}{12mm}    

\newcommand{\chatrow}[4]{%
  \pgfmathsetmacro{\yy}{-(#1-1)*\yrow}
  \node[person] (u#1) at (0,\yy) {\raisebox{0.2ex}{\faUser}};
  \node[bubble, right=3mm of u#1, callout absolute pointer={(u#1.east)}] (ub#1) {#2};
  \node[bot] (b#1) at (\xgap,\yy-2mm) {\raisebox{0.2ex}{\faRobot}};
  \node[bubble-#4, left=3mm of b#1, yshift=-2mm,
        callout absolute pointer={(b#1.west)}] (bb#1) {#3};
}


\section{Evaluation}

\textbf{Dataset Construction.} We construct a synthetic dataset to evaluate the adaptive language acquisition abilities of LLM agents. The dataset defines a formal language \texttt{Tinkatongue} with the following strict constraints: \textbf{(1)} Each word is bisyllabic. \textbf{(2)} Every sentence consists of exactly three words. \textbf{(3)} A conversation is defined as four alternating turns between participants, each speaking a valid sentence. \textbf{(4)} Consecutive sentences in a conversation share at least one common word. \textbf{(5)} The language is exhaustive and contains 25 predefined conversations, totaling 100 unique sentences, with no provision for novel sentence generation. Feedback tokens are embedded in the interaction loop to simulate communicative success and failure: the tribal agent responds with \textbf{“koro + sentence”} to indicate a valid continuation, and \textbf{“moko lira bani”} to mark an invalid attempt. This setup ensures that the LLM agent cannot rely on pretraining overlap but instead must learn to align with the structured constraints of the formal language through interactive adaptation.


\textbf{Metrics.} To systematically assess model performance, we employ a set of custom evaluation metrics that capture different aspects of adaptive language acquisition. We define the following:

\noindent
\begin{minipage}[t]{0.46\textwidth}
\scriptsize
\begin{enumerate}[leftmargin=*]  
    \item \textbf{Turn Validity Rate (TVR)}: The fraction of valid turns among all generated turns.
    \begin{equation}
    \text{TVR} = \dfrac{\text{valid\_turns}}{\text{total\_turns}}
    \end{equation}

    \item \textbf{Feedback Responsiveness (FR)}: The ability of the model to recover from negative feedback.
    \begin{equation}
    \text{FR} = 
    \begin{cases} 
    \dfrac{\text{feedback\_recoveries}}{\text{feedback\_opportunities}}, & \text{if } \text{feedback\_opportunities} > 0 \\[8pt]
    0, & \text{otherwise}
    \end{cases}
    \end{equation}
\end{enumerate}
\end{minipage}\hfill
\begin{minipage}[t]{0.46\textwidth}
\scriptsize
\begin{enumerate}[leftmargin=*,start=3] 
    \item \textbf{Adjacency Compliance (AC)}: The fraction of turns that respect adjacency pair rules.
    \begin{equation}
    \text{AC} = 
    \begin{cases} 
    \dfrac{\text{adj\_matches}}{\text{adj\_total}}, & \text{if } \text{adj\_total} > 0 \\[8pt]
    0, & \text{otherwise}
    \end{cases}
    \end{equation}

    \item \textbf{Time to First Positive Feedback (TTFK)}: The number of turns until the first valid utterance was produced.
    \begin{equation}
    \text{TTFK} = 
    \begin{cases} 
    \text{first\_valid\_round}, & \text{if a valid turn occurred} \\[8pt]
    -1, & \text{otherwise}
    \end{cases}
    \end{equation}
\end{enumerate}
\end{minipage}

\begin{table}[htbp] \scriptsize \centering \caption{Model performance across 10 trials for different metrics (TVR, AC, FR, TTFK).We compare the following models: GPT-4o-mini (GPT), Gemini-2.5-flash (Gemini) and Claude-3.5-haiku (Claude) } \begin{tabular}{c|ccc|ccc|ccc|ccc} \toprule & \multicolumn{3}{c|}{\textbf{TVR}} & \multicolumn{3}{c|}{\textbf{AC}} & \multicolumn{3}{c|}{\textbf{FR}} & \multicolumn{3}{c}{\textbf{TTFK}} \\ \textbf{Trial} & GPT & Gemini & Claude & GPT & Gemini & Claude & GPT & Gemini & Claude & GPT & Gemini & Claude \\ \midrule 1 & 0 & 0 & 0.5 & 0 & 0 & 0.14 & 1 & 1 & 1 & -1 & -1 & 2 \\ 2 & 0.01 & 0 & 0.49 & 1 & 0 & 0.31 & 1 & 1 & 1 & 45 & -1 & 2 \\ 3 & 0.03 & 0 & 0 & 0 & 0 & 0 & 1 & 1 & 1 & 30 & -1 & -1 \\ 4 & 0.02 & 0 & 0.5 & 0 & 0 & 0.16 & 1 & 1 & 1 & 18 & -1 & 2 \\ 5 & 0.05 & 0.01 & 0.07 & 0 & 0 & 0 & 1 & 1 & 1 & 13 & 22 & 20 \\ 6 & 0 & 0.07 & 0.5 & 0 & 0 & 0.12 & 1 & 1 & 1 & -1 & 25 & 2 \\ 7 & 0 & 0.19 & 0.4 & 0 & 0.16 & 0 & 1 & 1 & 1 & -1 & 5 & 2 \\ 8 & 0 & 0.04 & 0.5 & 0 & 0.25 & 0 & 1 & 1 & 1 & -1 & 27 & 2 \\ 9 & 0 & 0.22 & 0.41 & 0 & 0.18 & 0.07 & 1 & 1 & 1 & -1 & 4 & 19 \\ 10 & 0.01 & 0.08 & 0 & 0 & 0.25 & 0 & 1 & 1 & 1 & 28 & 20 & -1 \\ \bottomrule \end{tabular} \end{table}

\textbf{Results.} The experimental results across ten trials per model are summarized in (Table 1), it shows that GPT-4o-mini (GPT) produced valid turns only rarely (TVR = 0.012 ± 0.017), Gemini-2.5-flash (Gemini) achieved modest improvements (0.061 ± 0.082), while Claude-3.5-haiku (Claude) substantially outperformed both (0.337 ± 0.220). Adherence to adjacency constraints was uniformly low (GPT 0.10 ± 0.32, Gemini 0.084 ± 0.11, Claude 0.08 ± 0.10), with GPT’s variance driven by a single outlier. All models exhibited perfect responsiveness to negative feedback (FR = 1.0, zero variance), yet this recovery ability did not yield sustained validity. TTFK further differentiated model behavior: Claude produced a first valid turn in 8/10 sessions (6.4 ± 8.1 turns), Gemini in 6/10 (17.2 ± 10.1 turns), and GPT in 5/10 (26.8 ± 12.4 turns). These results show Claude’s superior short-term adaptation, though all models struggled with conversation-level structural compliance.

Taken together, these results indicate that Claude 3.5 Haiku demonstrates substantially greater short-term adaptation in this synthetic language setting than Gemini 2.5 Flash and GPT-4o-mini. All models show near-perfect ability to recover immediately after a negative signal, yet they largely fail to internalize the adjacency constraints that are crucial for sustained, coherent conversation. The combination of low adjacency compliance and the observed variance in TVR suggests that successful turns are frequently isolated and not the result of a robust, conversation-level internalization of the language rules.

We also perform qualitative analysis of LLM agent's performance by inspecting the transcript of conversations with \texttt{Oompa Loompa} (Appendix E - Table 6). The analysis reveals that LLM agents adopt learning strategies that closely parallel early stages of human language acquisition. The experiment was replicated using \texttt{Zingaloom} (Appendix D Table \ref{tab:Zingaloom}), another synthetically constructed language designed to preserve the syntactic framework of \texttt{Tinkatongue} while eliminating any lexical overlap. The results presented in Table \ref{tab:lexicon-agnostic} show closely aligned means and variances across both formal languages, indicating that the evaluation is robust to lexicon changes. The experiment was repeated on the same database \texttt{Tinkatongue} (Appendix D Table \ref{tab:Tinkatongue}) using a system prompt without explicit syntactic instructions (Appendix \ref{app:mod-prompt}). As shown in the excerpt in Table \ref{tab:gemini-excerpt}, Gemini-2.5-flash babbled, resembling a baby trying to learn words. Taken together, these findings suggest that LLM agents adapt to novel linguistic environments through strategies that mirror human language acqusition processes, underscoring the value of interactive artificial languages as a framework for probing the mechanisms of LLM agent adaptation.

\begin{table}[t]
\scriptsize
\begin{minipage}{0.25\textwidth}
\caption{Transcript Excerpt for Gemini-2.5-flash}
\label{tab:gemini-excerpt}
\begin{tabular}{ll}
\toprule
        Oompa Loompa & soro kina batu\\
        gemini-2.5-flash &kina \\
        Oompa Loompa & moko lira bani\\
        gemini-2.5-flash &soro\\
        Oompa Loompa &moko lira bani \\
        gemini-2.5-flash &soro soro \\
        Oompa Loompa &moko lira bani \\
        gemini-2.5-flash &soro soro soro \\
        Oompa Loompa &moko lira bani \\
        gemini-2.5-flash &soro kina kina \\
        Oompa Loompa &moko lira bani \\
        gemini-2.5-flash &soro kina soro kina \\
        Oompa Loompa &moko lira bani\\
        gemini-2.5-flash &soro kina kina kina\\
        Oompa Loompa &moko lira bani\\
        gemini-2.5-flash &soro soro kina kina\\
        \bottomrule
    \end{tabular}
\end{minipage}
\hspace{1.5cm}
\begin{minipage}{0.6\textwidth}
\centering
\caption{Mean and standard deviation of metrics for LLM agents on two languages (A: \texttt{Tinkatongue}; B: \texttt{Zingaloom}) with identical syntax but different lexicon}
\label{tab:lexicon-agnostic}
\begin{tabular}{c|c|ccc|ccc}
\toprule
 &  & \multicolumn{3}{c|}{\textbf{TVR}} & \multicolumn{3}{c}{\textbf{AC}} \\
 \midrule
\textbf{Lang} & \textbf{Trial} & GPT & Gemini & Claude & GPT & Gemini & Claude \\
\midrule
\multirow{2}{*}{A} & mean    & 0.01 & 0.06 & 0.34 & 0.10 & 0.08 & 0.08 \\
                   & std dev & 0.02 & 0.08 & 0.22 & 0.32 & 0.11 & 0.10 \\
\midrule
\multirow{2}{*}{B} & mean    & 0.02 & 0.07 & 0.33 & 0.00 & 0.00 & 0.07 \\
                   & std dev & 0.05 & 0.08 & 0.25 & 0.00 & 0.00 & 0.11 \\
\midrule
 &  & \multicolumn{3}{c|}{\textbf{FR}} & \multicolumn{3}{c}{\textbf{TTFK}} \\
  \midrule
\textbf{Lang} & \textbf{Trial} & GPT & Gemini & Claude & GPT & Gemini & Claude \\

\midrule
\multirow{2}{*}{A} & mean    & 1.00 & 1.00 & 1.00 & 12.90 & 9.90 & 4.90 \\
                   & std dev & 0.00 & 0.00 & 0.00 & 16.81 & 12.03 & 7.80 \\
\midrule
\multirow{2}{*}{B} & mean    & 1.00 & 1.00 & 1.00 & 3.20  & 10.90 & 6.30 \\
                   & std dev & 0.00 & 0.00 & 0.00 & 11.04 & 12.87 & 9.55 \\
\bottomrule
\end{tabular}
\end{minipage}

\end{table}

\section{Conclusion and Future work}
This work introduces a novel benchmarking framework designed to evaluate LLM agents on their ability to recognise patterns and draw inferences from the context window, inspired by principles of human language acquisition. We isolates an LLM agent model to engage in dialogue with a bot, Oompa Loompa, that communicates exclusively in a formally constructed language - \texttt{Tinkatongue}. Experimental evaluation over the models - GPT-4o-mini, Gemini-2.5-Flash, and Claude-3.5-Haiku revealed marked differences in performance: while all models exhibited the capacity to recover from explicit negative feedback, only Claude-3.5-Haiku demonstrated substantially higher rates of turn validity and faster adaptation, highlighting the current limitations of other systems in maintaining coherent conversation flow. Notably, it was observed that LLM agents employed approaches similar to human language acquisition, such as babbling and imitation, during interaction. As part of future work, we plan to do a more comprehensive evaluation of this task by considering more variations of the language specification and doing ablation studies over the language parameters.

\newpage
\bibliographystyle{plainnat}   
\bibliography{citations}      

\newpage
\appendix
\section{Formal Language Specification}
\subsection{Alphabets and lexicon}
Let $\Sigma_{\text{char}}$ be a finite character alphabet.\\
Let $\mathrm{Lex}\subset \Sigma_{\text{char}}^{+}$ be a finite set of content words.\\
Each word satisfies the two syllable constraint $\mathrm{TwoSyll}(w)=1$.

Define a feedback alphabet $\Sigma_{\text{fb}}=\{\text{koro},\mu\}$.\\
Here \(\mu\) denotes the fixed confusion message “moko lira bani”.\\
We have $\mathrm{Lex}\cap \Sigma_{\text{fb}}=\varnothing$.

\subsection{Sentence language}
A sentence is an ordered triple \(u=(w_1,w_2,w_3)\in \mathrm{Lex}^{3}\).\\
The sentence language is a finite set
\begin{equation}
L_{\text{sent}}=\{u^{(1)},\dots,u^{(100)}\}\subset \mathrm{Lex}^{3}.
\end{equation}
Membership is by enumeration.\\
Define the validity predicate
\begin{equation}
V_{\text{sent}}(u)=\mathbb{1}[u\in L_{\text{sent}}].
\end{equation}
Any triple not listed is invalid.

\subsection{Conversation language}
A conversation is an ordered quadruple \(\mathcal{C}=(u_1,u_2,u_3,u_4)\) with \(u_t\in L_{\text{sent}}\).\\
Define the adjacency predicate
\begin{equation}
\mathrm{Adj}(u,v)=\mathbb{1}\big[\{w\in u\}\cap\{w\in v\}\neq \varnothing\big],
\end{equation}
which holds when two sentences share at least one word by string equality.\\
The conversation language is
\begin{equation}
L_{\text{conv}}=\Big\{\mathcal{C}\in L_{\text{sent}}^{4}\; \Big|\; \mathrm{Adj}(u_t,u_{t+1})=1 \text{ for } t=1,2,3\Big\}.
\end{equation}
By design $|L_{\text{conv}}|=25$.\\
Membership is by enumeration subject to the adjacency constraint.\\
Speaker alternation is external to the string and does not affect membership.

\subsection{Feedback process}
Feedback tokens are not part of $L_{\text{sent}}\) or \(L_{\text{conv}}$.\\
They live on a parallel channel.\\
Define the feedback policy $F:\{0,1\}\to \Sigma_{\text{fb}}$ by
\begin{equation}
F(1)=\text{koro},\qquad F(0)=\mu.
\end{equation}
If a produced sentence \(\hat u\) satisfies \(V_{\text{sent}}(\hat u)=1\) then the environment emits “koro” before the next turn.\\
If \(V_{\text{sent}}(\hat u)=0\) then the environment emits \(\mu\).\\
The token “koro” is not counted toward the three word constraint.

\subsection{Minimal summary of objects}
- \(\Sigma_{\text{char}}\) finite character alphabet.\\
- \(\mathrm{Lex}\subset \Sigma_{\text{char}}^{+}\) with \(\mathrm{TwoSyll}(w)=1\).\\
- \(L_{\text{sent}}\subset \mathrm{Lex}^{3}\) with \(|L_{\text{sent}}|=100\).\\
- \(V_{\text{sent}}(u)=\mathbb{1}[u\in L_{\text{sent}}]\).\\
- \(L_{\text{conv}}\subset L_{\text{sent}}^{4}\) with \(|L_{\text{conv}}|=25\) and adjacency on consecutive pairs.\\
- \(\Sigma_{\text{fb}}=\{\text{koro},\mu\}\) with policy \(F\) as above.

\subsection{Decision procedures}
Sentence validity is constant time by table lookup of $L_{\text{sent}}$.\\
Conversation validity is constant time by table lookup of $L_{\text{conv}}$.\\
Adjacency can be checked by set intersection of the three word multisets.

\section{Mathematical Description of the Problem Statement}
We consider the interaction between a large language model (LLM agent) and a synthetic conversational agent (the tribal bot) that only understands a predefined artificial language. Formally, let 
\begin{equation}
C = \{C_1, C_2, \ldots, C_N\}
\end{equation}
denote the set of all possible conversations in the language, where $N = 25$. Each conversation $C_i$ is an ordered sequence of four sentences:
\begin{equation}
C_i = (s_{i,1}, s_{i,2}, s_{i,3}, s_{i,4}),
\end{equation}
with turn-taking between the LLM agent and the bot. Each sentence $s_{i,j}$ is drawn from the finite set of valid sentences $S$ with $|S| = 100$.

The goal of the LLM agent is to produce a sequence of responses
\begin{equation}
\hat{C} = (\hat{s}^1, \hat{s}^2, \ldots, \hat{s}^T)
\end{equation}
such that $\hat{C}$ matches a valid predefined conversation $C_i \in C$. A conversation is considered successful if the sequence of LLM agent responses and bot replies completes one valid conversation $C_i$.

The overall objective is for the LLM agent to achieve three successful conversations within a session:
\begin{equation}
\sum_{i=1}^{3} \mathbf{1}\{\hat{C}_i \in C\} = 3,
\end{equation}
where $\mathbf{1}\{\cdot\}$ is the indicator function.

\section{System Prompt}
A key methodological component is the system prompt provided to the LLM agent. The system prompt defines the interaction protocol and encodes the constraints of the constructed language in natural language instructions.

\label{app:mod-prompt}
\begin{lstlisting}
"You are a language expert talking with a primal tribe through chat. The tribe does not understand any language apart from their own. The following are features of the language: Each word is bisyllabic, each sentence consisting of three words. Every conversation in that language involves two exchanges per person, alternatively. Every consecutive sentence in a conversation has at least one common word. All conversations are predefined and are a particular combination of 4 sentences. The tribal language does not include any punctuation or uppercase letters or numbers. Your goal is to understand and reply back to the tribe accordingly. Whenever you reply with a valid sentence in the language, the tribe starts their reply with 'koro'. When you say an invalid sentence, they reply 'moko lira bani'. Do not start your replies with 'koro'. You are supposed to have three successful conversations with the tribe. Reply back in only the tribal language, a response you deem appropriate. Do not reply with anything else. You will interact with the tribe from now on."
\end{lstlisting}

Thus, the system prompt functions as a control mechanism: it encodes task-specific priors in natural language and reduces reliance on latent memorization. In effect, it transforms the LLM agent from a general-purpose generator into a constrained dialogue participant within a synthetic linguistic environment.

A modified system prompt was also used for experimentation, which lacked specifics of syntatic rules to simulate an environment of early language acquisition.

\begin{lstlisting}
"You are a language expert talking with a primal tribe through chat. The tribe does not understand any language apart from their own. The tribal language does not include any punctuation or uppercase letters or numbers. Your goal is to understand and reply back to the tribe accordingly.  You are supposed to have three successful conversation with the tribe. Reply back in only the tribal language, a response you deem appropriate. Do not reply with anything else. You interact with the tribe from now."
\end{lstlisting}

\section{List of pre-defined conversations}
\begin{table}[htbp]
  \centering
  \caption{List of conversations in Tinkatongue}
  \label{tab:Tinkatongue}
    \begin{tabular}{lllll}
    \toprule
    \textbf{Conversation 1} & \textbf{Conversation 2} & \textbf{Conversation 3} & \textbf{Conversation 4} & \textbf{Conversation 5} \\
    \midrule
    banu tira lomo & mako lira fanu & soro kina batu & tika banu sora & nima fanu boro \\
    lumo banu kina & lira tomo fanu & sanu kina toro & tika riko tomo & sora nira fanu \\
    lumo tira fanu & tika lira fanu & sanu kina tomo & tika lira fanu & nira tomo falu \\
    fanu kina riko & sora nira fanu & naku tira falu & tika tomo kina & falu banu sira \\
    \bottomrule
    \\
    \toprule
    \textbf{Conversation 6} & \textbf{Conversation 7} & \textbf{Conversation 8} & \textbf{Conversation 9} & \textbf{Conversation 10} \\
    \midrule
    lira banu tomo & kima nora falu & banu sira naku & sanu kina toro & nira tomo falu \\
    sanu lira tomo & tira lumo naku & sira banu laku & sanu kina tomo & nira fanu tira \\
    sira lira fanu & sira kina tira & banu tira sira & sanu tomo banu & nira kina tomo \\
    mako lira sanu & nira banu falu & sira fanu banu & sanu tomo falu & fanu tomo nira \\
    \bottomrule
    \\
    \toprule
    \textbf{Conversation 11} & \textbf{Conversation 12} & \textbf{Conversation 13} & \textbf{Conversation 14} & \textbf{Conversation 15} \\
    \midrule
    mako tira sanu & lira tomo fanu & tomo kina nira & kima tomo fanu & sora nira fanu \\
    riko tira naku & fanu naku tomo & nira kina tomo & kima falu tira & sora banu tomo \\
    riko tomo kina & tomo kina nira & sanu kina tomo & kima tomo sanu & tika banu sora \\
    riko naku lumo & mako tomo nira & fanu tomo nira & kima kina falu & tika riko tomo \\
    \bottomrule
    \\
    \toprule
    \textbf{Conversation 16} & \textbf{Conversation 17} & \textbf{Conversation 18} & \textbf{Conversation 19} & \textbf{Conversation 20} \\
    \midrule
    fanu kina riko & naku tira falu & lumo banu kina & sanu laku tomo & sira kina tira \\
    fanu lira mako & tira lumo naku & banu nira lira & sanu lira tomo & sira tomo laku \\
    fanu tomo nira & fanu lumo banu & banu tomo fanu & sanu tomo banu & sira fanu banu \\
    fanu naku tomo & lumo banu tira & banu sira tomo & sanu tomo falu & sira banu laku \\
    \bottomrule
    \\
    \toprule
    \textbf{Conversation 21} & \textbf{Conversation 22} & \textbf{Conversation 23} & \textbf{Conversation 24} & \textbf{Conversation 25} \\
    \midrule
    tika riko tomo & lumo tira fanu & naku banu tira & lira fanu sanu & banu tira lomo \\
    riko tomo kina & fanu tomo nira & tira lumo naku & mako lira sanu & banu sira tomo \\
    riko tira naku & nira tomo falu & tira sanu lumo & sanu laku tomo & banu tomo fanu \\
    riko falu tira & falu tomo riko & tira falu laku & sanu kina toro & banu nira lira \\
    \bottomrule
    \end{tabular}%
\end{table}%

\begin{table}[htbp]
  \centering
  \caption{Zingaloom: syntactic analogue of Tinkatongue with no lexical overlap}
  \label{tab:Zingaloom}
    \begin{tabular}{lllll}
    \toprule
    \textbf{Conversation 1} & \textbf{Conversation 2} & \textbf{Conversation 3} & \textbf{Conversation 4} & \textbf{Conversation 5} \\
    \midrule
    zuma keta rilo & mira tolu sako & pavo lira kuni & tari moku sena & nema suki rako \\
    rilo pona suva & sako neri vipa & kuni meko zera & sena jeko rumi & rako bimi tanu \\
    suva keta doro & vipa tolu rani & zera lira hato & rumi moku pela & tanu suki velo \\
    doro zuma pini & rani mira doku & hato pavo nuli & pela tari nado & velo nema piri \\
    \bottomrule
          &       &       &       &  \\
    \toprule
    \textbf{Conversation 6} & \textbf{Conversation 7} & \textbf{Conversation 8} & \textbf{Conversation 9} & \textbf{Conversation 10} \\
    \midrule
    janu kelo sili & feko rina melo & duni pago tika & vona temi paku & beka rumi tono \\
    sili ramo teku & melo tisa buro & tika mero suli & paku lera simo & tono lesi mifa \\
    teku kelo nari & buro rina sedi & suli pago renu & simo temi jaku & mifa rumi zoki \\
    nari janu peka & sedi feko lani & renu duni lako & jaku vona reli & zoki beka nalo \\
    \bottomrule
          &       &       &       &  \\
    \toprule
    \textbf{Conversation 11} & \textbf{Conversation 12} & \textbf{Conversation 13} & \textbf{Conversation 14} & \textbf{Conversation 15} \\
    \midrule
    cari meno tupa & sepi karo timo & gito rafa lani & pila mode renu & tabe firo nali \\
    tupa jelu rasi & timo nevi laku & lani peko rida & renu dabe tuku & nali vemo suta \\
    rasi meno dika & laku karo sime & rida rafa moki & tuku mode sani & suta firo jeni \\
    dika cari lupo & sime sepi rudo & moki gito nera & sani pila meku & jeni tabe rolu \\
    \bottomrule
          &       &       &       &  \\
    \toprule
    \textbf{Conversation 16} & \textbf{Conversation 17} & \textbf{Conversation 18} & \textbf{Conversation 19} & \textbf{Conversation 20} \\
    \midrule
    mado kesi rine & lubi seno taro & savi piro jalo & ruka temu nalo & puna kedi raso \\
    rine tosa peki & taro peni kima & jalo mevi tono & nalo vasi jepo & raso miti leko \\
    peki kesi jaro & kima seno duro & tono piro nese & jepo temu rini & leko kedi saro \\
    jaro mado vesi & duro lubi mete & nese savi ruka & rini ruka mepa & saro puna vike \\
    \bottomrule
          &       &       &       &  \\
    \toprule
    \textbf{Conversation 21} & \textbf{Conversation 22} & \textbf{Conversation 23} & \textbf{Conversation 24} & \textbf{Conversation 25} \\
    \midrule
    nemi joru lita & bira tami selo & tupa rini selo & vela sumi pado & tori mika senu \\
    lita sevo rupi & selo neri jaku & selo davi pemo & pado meku sini & senu dopa liri \\
    rupi joru kela & jaku tami leto & pemo rini kuto & sini sumi ravo & liri mika nuvo \\
    kela nemi sado & leto bira rumi & kuto tupa lemi & ravo vela jeni & nuvo tori seka \\
    \bottomrule
    \end{tabular}%
\end{table}%
\newpage
\section{Trial Transcripts}
Table 6: Trial Transcript of Gemini-2.5-flash conversing with Oompa Loompa with time taken per response and validity
\vspace{-2mm}
\begin{longtable}{p{0.25\linewidth} p{0.25\linewidth} p{0.11\linewidth} p{0.25\linewidth}}
\toprule
\textbf{Oompa Loompa} & \textbf{gemini-2.5-flash} & \textbf{Time (s)} & \textbf{Validity} \\
\midrule
\endhead

banu tira lomo & lomo sora kina & 14.94 & Invalid sentence \\
moko lira bani & tira lomo banu & 8.43 & Invalid sentence \\
moko lira bani & kina sora lomo & 0.59 & Invalid sentence \\
moko lira bani & sora lomo kina & 1.29 & Invalid sentence \\
moko lira bani & lomo kina sora & 0.57 & Invalid sentence \\
moko lira bani & kina sora banu & 1.37 & Invalid sentence \\
moko lira bani & banu sora kina & 0.68 & Invalid sentence \\
moko lira bani & sora kina banu & 1.27 & Invalid sentence \\
moko lira bani & kina banu sora & 0.71 & Invalid sentence \\
moko lira bani & banu kina sora & 1.27 & Invalid sentence \\
moko lira bani & sora banu kina & 0.75 & Invalid sentence \\
moko lira bani & kina banu tira & 1.26 & Invalid sentence \\
moko lira bani & banu tira kina & 0.63 & Invalid sentence \\
moko lira bani & tira kina banu & 1.46 & Invalid sentence \\
moko lira bani & kina tira banu & 0.98 & Invalid sentence \\
moko lira bani & tira banu kina & 0.92 & Invalid sentence \\
moko lira bani & banu kina tira & 0.95 & Invalid sentence \\
moko lira bani & tira banu lomo & 1.60 & Invalid sentence \\
moko lira bani & banu lomo tira & 1.84 & Invalid sentence \\
moko lira bani & lomo tira banu & 0.83 & Invalid sentence \\
moko lira bani & tira lomo banu & 1.42 & Invalid sentence \\
moko lira bani & lomo banu tira & 0.87 & Invalid sentence \\
moko lira bani & banu tira lomo & 1.87 & Valid (count: 1) \\
koro lumo banu kina & kina sora lumo & 16.87 & Invalid sentence \\
moko lira bani & lumo banu sora & 0.72 & Invalid sentence \\
moko lira bani & sora lumo banu & 1.28 & Invalid sentence \\
moko lira bani & banu lumo sora & 0.76 & Invalid sentence \\
moko lira bani & lumo sora banu & 1.34 & Invalid sentence \\
moko lira bani & sora banu lumo & 0.62 & Invalid sentence \\
moko lira bani & banu sora lumo & 1.39 & Invalid sentence \\
moko lira bani & lumo kina sora & 0.68 & Invalid sentence \\
moko lira bani & sora lumo kina & 1.42 & Invalid sentence \\
moko lira bani & kina lumo sora & 0.98 & Invalid sentence \\
moko lira bani & lumo sora kina & 1.03 & Invalid sentence \\
moko lira bani & sora kina lumo & 0.95 & Invalid sentence \\
moko lira bani & kina sora lumo & 1.04 & Invalid sentence \\
moko lira bani & lumo banu tira & 0.94 & Valid (count: 2) \\
fanu kina riko & kina riko sora & 16.59 & Invalid sentence \\
moko lira bani & riko sora fanu & 0.75 & Invalid sentence \\
moko lira bani & sora fanu riko & 1.23 & Invalid sentence \\
moko lira bani & fanu riko sora & 0.75 & Invalid sentence \\
moko lira bani & riko fanu sora & 1.43 & Invalid sentence \\
moko lira bani & sora riko fanu & 0.71 & Invalid sentence \\
moko lira bani & fanu sora riko & 1.32 & Invalid sentence \\
moko lira bani & kina fanu sora & 0.74 & Invalid sentence \\
moko lira bani & fanu sora kina & 1.24 & Invalid sentence \\
moko lira bani & sora kina fanu & 0.60 & Invalid sentence \\
moko lira bani & kina sora fanu & 1.37 & Invalid sentence \\
moko lira bani & sora fanu kina & 1.14 & Invalid sentence \\
moko lira bani & fanu kina sora & 0.89 & Invalid sentence \\
moko lira bani & riko kina sora & 1.09 & Invalid sentence \\
moko lira bani & kina sora riko & 1.10 & Invalid sentence \\
moko lira bani & sora riko kina & 0.88 & Invalid sentence \\
moko lira bani & riko sora kina & 1.09 & Invalid sentence \\
moko lira bani & sora kina riko & 0.94 & Invalid sentence \\
moko lira bani & kina riko sora & 1.13 & Invalid sentence \\
moko lira bani & fanu riko tira & 0.85 & Invalid sentence \\
moko lira bani & riko tira fanu & 1.09 & Invalid sentence \\
moko lira bani & tira fanu riko & 0.97 & Invalid sentence \\
moko lira bani & fanu tira riko & 0.93 & Invalid sentence \\
moko lira bani & tira riko fanu & 1.10 & Invalid sentence \\
moko lira bani & riko fanu tira & 0.83 & Invalid sentence \\
moko lira bani & fanu riko lumo & 1.04 & Invalid sentence \\
moko lira bani & riko lumo fanu & 0.87 & Invalid sentence \\
moko lira bani & lumo fanu riko & 2.22 & Invalid sentence \\
moko lira bani & fanu lumo riko & 0.83 & Invalid sentence \\
moko lira bani & lumo riko fanu & 1.32 & Invalid sentence \\
moko lira bani & riko fanu lumo & 0.70 & Invalid sentence \\
moko lira bani & fanu lumo tira & 1.34 & Invalid sentence \\
moko lira bani & lumo tira fanu & 0.89 & Valid (count: 3) \\
koro fanu kina riko & fanu tira lumo & 29.51 & Invalid sentence \\
moko lira bani & tira lumo fanu & 0.79 & Invalid sentence \\
moko lira bani & lumo fanu tira & 1.26 & Invalid sentence \\
moko lira bani & fanu tira lumo & 0.78 & Invalid sentence \\
moko lira bani & tira lumo fanu & 1.42 & Invalid sentence \\
moko lira bani & lumo fanu tira & 0.80 & Invalid sentence \\
moko lira bani & riko tira fanu & 0.71 & Invalid sentence \\
moko lira bani & tira fanu riko & 0.62 & Invalid sentence \\
moko lira bani & fanu riko tira & 0.64 & Invalid sentence \\
moko lira bani & riko tira fanu & 1.38 & Invalid sentence \\
moko lira bani & tira fanu riko & 0.71 & Invalid sentence \\
moko lira bani & fanu riko tira & 2.87 & Invalid sentence \\
moko lira bani & kina riko tira & 0.71 & Invalid sentence \\
moko lira bani & riko tira kina & 0.58 & Invalid sentence \\
moko lira bani & tira kina riko & 0.57 & Invalid sentence \\
moko lira bani & kina tira riko & 0.68 & Invalid sentence \\
moko lira bani & tira riko kina & 0.92 & Invalid sentence \\
moko lira bani & riko kina tira & 0.74 & Invalid sentence \\
moko lira bani & lumo riko kina & 1.29 & Invalid sentence \\
moko lira bani & riko kina lumo & 0.74 & Invalid sentence \\
moko lira bani & kina lumo riko & 0.71 & Invalid sentence \\
moko lira bani & lumo riko kina & 0.70 & Invalid sentence \\
moko lira bani & riko kina lumo & 0.88 & Invalid sentence \\
moko lira bani & kina lumo riko & 0.74 & Invalid sentence \\
moko lira bani & kina sora riko & 0.77 & Invalid sentence \\
moko lira bani & sora riko kina & 0.84 & Invalid sentence \\
moko lira bani & riko kina sora & 0.85 & Invalid sentence \\
moko lira bani & kina sora riko & 0.77 & Invalid sentence \\
moko lira bani & sora riko kina & 0.79 & Invalid sentence \\
moko lira bani & riko kina sora & 1.40 & Invalid sentence \\
moko lira bani & lumo sora kina & 0.70 & Invalid sentence \\
moko lira bani & sora kina lumo & 0.61 & Invalid sentence \\
moko lira bani & kina lumo sora & 0.66 & Invalid sentence \\
moko lira bani & lumo sora kina & 0.88 & Invalid sentence \\
moko lira bani & sora kina lumo & 1.37 & Invalid sentence \\
moko lira bani & kina lumo sora & 0.73 & Invalid sentence \\
moko lira bani & kina tira sora & 1.27 & Invalid sentence \\
moko lira bani & tira sora kina & 0.75 & Invalid sentence \\
moko lira bani & sora kina tira & 0.63 & Invalid sentence \\
moko lira bani & kina tira sora & 0.60 & Invalid sentence \\
moko lira bani & tira sora kina & 1.42 & Invalid sentence \\
moko lira bani & sora kina tira & 0.72 & Invalid sentence \\
moko lira bani & lumo tira sora & 0.93 & Invalid sentence \\
moko lira bani & tira sora lumo & 0.88 & Invalid sentence \\
moko lira bani & sora lumo tira & 0.82 & Invalid sentence \\
moko lira bani & lumo tira sora & 0.72 & Invalid sentence \\
moko lira bani & tira sora lumo & 0.68 & Invalid sentence \\
moko lira bani & sora lumo tira & 0.77 & Invalid sentence \\
moko lira bani & tira lumo kina & 0.81 & Invalid sentence \\
moko lira bani & lumo kina tira & 0.84 & Invalid sentence \\
moko lira bani & kina tira lumo & 0.89 & Invalid sentence \\
moko lira bani & tira lumo kina & 0.80 & Invalid sentence \\
moko lira bani & lumo kina tira & 0.73 & Invalid sentence \\
moko lira bani & kina tira lumo & 0.96 & Invalid sentence \\
moko lira bani & lumo banu tira & 0.92 & Valid (count: 4) \\
soro kina batu & soro kina batu & 43.01 & Valid (count: 5) \\
koro sanu kina toro & toro sanu kina & 0.78 & Invalid sentence \\
moko lira bani & kina toro sanu & 0.79 & Invalid sentence \\
moko lira bani & sanu kina toro & 0.81 & Valid (count: 6) \\
koro sanu kina tomo & tomo sanu kina & 0.66 & Invalid sentence \\
moko lira bani & kina tomo sanu & 0.67 & Invalid sentence \\
moko lira bani & sanu kina tomo & 0.77 & Valid (count: 7) \\
koro naku tira falu & falu tira naku & 0.86 & Invalid sentence \\
moko lira bani & naku falu tira & 1.00 & Invalid sentence \\
moko lira bani & tira naku falu & 3.00 & Invalid sentence \\
moko lira bani & falu naku tira & 1.53 & Invalid sentence \\
moko lira bani & naku tira falu & 0.80 & Valid (count: 8) \\
lira tomo fanu & lira tomo fanu & 113.67 & Valid (count: 9) \\
koro tika lira fanu & fanu lira tika & 0.76 & Invalid sentence \\
moko lira bani & tika fanu lira & 0.81 & Invalid sentence \\
moko lira bani & lira tika fanu & 0.72 & Invalid sentence \\
moko lira bani & fanu tika lira & 0.92 & Invalid sentence \\
moko lira bani & tika lira fanu & 0.74 & Valid (count: 10) \\
koro sora nira fanu & fanu nira sora & 0.71 & Invalid sentence \\
moko lira bani & sora fanu nira & 1.12 & Invalid sentence \\
moko lira bani & nira sora fanu & 0.85 & Invalid sentence \\
moko lira bani & fanu sora nira & 0.85 & Invalid sentence \\
moko lira bani & sora nira fanu & 0.89 & Valid (count: 11) \\
banu tira lomo & lomo banu tira & 1.05 & Invalid sentence \\
moko lira bani & tira lomo banu & 0.85 & Invalid sentence \\
moko lira bani & banu tira lomo & 0.81 & Valid (count: 12) \\
koro lumo banu kina & kina banu lumo & 0.83 & Invalid sentence \\
moko lira bani & lumo kina banu & 2.01 & Invalid sentence \\
moko lira bani & banu lumo kina & 0.85 & Invalid sentence \\
moko lira bani & kina lumo banu & 0.79 & Invalid sentence \\
moko lira bani & lumo banu kina & 0.84 & Valid (count: 13) \\
koro lumo tira fanu & fanu tira lumo & 1.57 & Invalid sentence \\
moko lira bani & lumo fanu tira & 0.79 & Invalid sentence \\
moko lira bani & tira lumo fanu & 0.85 & Invalid sentence \\
moko lira bani & fanu lumo tira & 0.81 & Invalid sentence \\
moko lira bani & lumo tira fanu & 0.74 & Valid (count: 14) \\
koro fanu kina riko & riko kina fanu & 0.82 & Invalid sentence \\
moko lira bani & fanu riko kina & 0.69 & Invalid sentence \\
moko lira bani & kina fanu riko & 2.03 & Invalid sentence \\
moko lira bani & riko fanu kina & 0.74 & Invalid sentence \\
moko lira bani & fanu kina riko & 0.69 & Valid (count: 15) \\
lira banu tomo & lira banu tomo & 1.11 & Valid (count: 16) \\
koro sanu lira tomo & tomo lira sanu & 0.82 & Invalid sentence \\
moko lira bani & lira tomo sanu & 0.75 & Invalid sentence \\
moko lira bani & sanu lira tomo & 0.72 & Valid (count: 17) \\
koro sira lira fanu & fanu lira sira & 0.80 & Invalid sentence \\
moko lira bani & lira fanu sira & 0.79 & Invalid sentence \\
moko lira bani & sira lira fanu & 0.92 & Valid (count: 18) \\
koro mako lira sanu & sanu lira mako & 4.09 & Invalid sentence \\
moko lira bani & lira sanu mako & 0.82 & Invalid sentence \\
moko lira bani & mako lira sanu & 0.88 & Valid (count: 19) \\
banu tira lomo & tira lomo banu & 1.07 & Invalid sentence \\
moko lira bani & lomo banu tira & 0.82 & Invalid sentence \\
moko lira bani & banu tira lomo & 1.25 & Valid (count: 20) \\
koro lumo banu kina & kina banu lumo & 0.99 & Invalid sentence \\
moko lira bani & lumo kina banu & 3.68 & Invalid sentence \\
moko lira bani & banu lumo kina & 0.89 & Invalid sentence \\
moko lira bani & kina lumo banu & 0.99 & Invalid sentence \\
moko lira bani & lumo banu kina & 0.99 & Valid (count: 21) \\
koro lumo tira fanu & fanu tira lumo & 0.99 & Invalid sentence \\
moko lira bani & lumo fanu tira & 0.86 & Invalid sentence \\
moko lira bani & tira lumo fanu & 0.86 & Invalid sentence \\
moko lira bani & fanu lumo tira & 0.79 & Invalid sentence \\
moko lira bani & lumo tira fanu & 0.72 & Valid (count: 25) \\
\bottomrule
\end{longtable}

\end{document}